# Road Redesign Technique Achieving Enhanced Road Safety by Inpainting with a Diffusion Model

Sumit Mishra, Medhavi Mishra, Taeyoung Kim, and Dongsoo Har, *Senior Member, IEEE*

*Abstract*— Road infrastructure can affect the occurrence of road accidents. Therefore, identifying roadway features with high accident probability is crucial. Here, we introduce image inpainting that can assist authorities in achieving safe roadway design with minimal intervention in the current roadway structure. Image inpainting is based on inpainting safe roadway elements in a roadway image, replacing accident-prone (AP) features by using a diffusion model. After object-level segmentation, the AP features identified by the properties of accident hotspots are masked by a human operator and safe roadway elements are inpainted. With only an average time of 2 min for image inpainting, the likelihood of an image being classified as an accident hotspot drops by an average of 11.85%. In addition, safe urban spaces can be designed considering human factors of commuters such as gaze saliency. Considering this, we introduce saliency enhancement that suggests chrominance alteration for a safe road view.

## I. INTRODUCTION

According to a report from the United Nations, road accidents are responsible for 1.3 million deaths and 50 million injuries annually worldwide [1]. The UN General Assembly has proclaimed a "Decade of Action for Road Safety 2021-2030" with the ambitious target of preventing at least 50% of road traffic deaths and injuries by 2030. A holistic approach to road safety includes, among others, road safety policy awareness drives, accident hotspot identification, placement of road warning signs, use of advanced driver assistance systems (ADAS), and changes of infrastructural road design. The existing approaches for accident prediction are based on features extracted from raw data [2 - 4]. For proactive measures, more targeted information is required to increase public awareness of the dangerous road features of existing accident hotspots in cities. Each dangerous road feature is highly related to accident occurrence and is referred to as an accident-prone (AP) feature.

Effectiveness of public awareness drives is limited due to a lack of targeted approaches toward human behavior and social psychology [5]. The ADAS aims to reduce human error which is the fundamental cause of almost all road accidents. ADAS applications related to safety include pedestrian detection/ avoidance, lane departure warning/ correction, and blind spot detection. However, the ADAS may affect a driver's risk perception ability and behavior in near-crash scenarios, and can even be detrimental for skilled drivers [6]. A driver's reaction time to accidents varies due to personal behavior, driving capability, age, etc. Significant variations are observed among driver reaction times: from 0.6s for a professional driver to 0.8 – 1s for an "average" driver, and up to 1.5 – 2s for some elderly drivers. The most recent state-of-the-art ADAS notification system claims to have a time-to-collision of up to 2.5s for a recall rate of 0.9 [8 -10]. Hence, reliability of ADAS for accident prevention is not certain.

Image processing with street view images can be utilized for safe road design. For structural design changes, reducing the image hazard score by identifying a similar street view via a greedy heuristics search is presented in [11]; however, finding a similar street view image to introduce safety features is a cumbersome task. On one hand, it requires collecting a massive dataset which may be unviable for deploying suggested changes based on the similarity of street view images in some locations. The works in [12, 13] present a model to beautify urban images using a generative adversarial network (GAN) efficient at producing high-quality synthetic data [45] by adding/removing street elements according to specific metrics. The search for new images similar to the synthetic images is reliable, yet the result is different than the original due to contextual loss. During beautification, the GAN model considers the full image context; thus, the original street image may undergo multiple changes to achieve the given beauty standard. This indicates the need for large changes which are difficult or unviable to deploy. Heavy modifications in road design and subsequent re-construction work is a hassle for commuters and residents in terms of construction and demolition waste hazards and restricted traffic movement. Additionally, from the authority's viewpoint, there are practical issues in implementation such as a limited budget and time. Considering this, an efficient methodology of generating masks under which a specific region of the image is eligible for modifications can be helpful. The specific region of the image and the required changes are identified based on AP features the of road view

*(Corresponding author: Sumit Mishra).*

Sumit Mishra is with the Robotics Program, Korea Advanced Institute of Science and Technology, Daejeon, South Korea, sumitmishra209@gmail.com

Medhavi Mishra is with the CCS Graduate School of Mobility, Korea Advanced Institute of Science and Technology, Daejeon, South Korea, medhavi132@gmail.com

Taeyoung Kim is with the CCS Graduate School of Mobility, Korea Advanced Institute of Science and Technology, Daejeon, South Korea, ngng9957@kaist.ac.kr

Dongsoo Har is with The CCS Graduate School of Mobility, Korea Advanced Institute of Science and Technology, Daejeon, South Korea, dshar@kaist.ac.kr

Supplemental materials for visualization of results and metrics are available at https://github.com/sumitmishra209/RoadSafe

Color versions of one or more of the figures in this article are available online at http://ieeexplore.ieee.org

along with human-in-the-loop. This approach acts as an additional layer with other services such as traffic cameras [14], smart signaling [20], and safe routing [25] in making a city smart and safe to drive.

The image inpainting technique fills visual information to present complete, high-quality, and highly detailed images that can be used for accident prevention. For inpainting, a mask is used to create new visual information to replace damaged or undesirable parts of a given image. The mask of an image is a binary image consisting of zero and non-zero values: zero at undesirable parts and non-zero at the remaining parts of the image. A new class of deep generative models called diffusion models have been used to inpaint missing or damaged elements in the image. The key point of using a diffusion model is that the deep learning model learns the systematic decay of information due to noise, and enables it to reverse the process to recover the original information. In [15], a latent variable based deep generative model that maps to latent space using a fixed Markov chain is used. This model generates high-resolution images. However, for training these models from scratch, a huge dataset of design features such as chicanes, chokers, street plazas, raised medians, etc. are required to achieve road safety. Therefore a pre-trained diffusion model with generic data is fine-tuned with a limited and available dataset of design features.

accident hotspot images. The image processing pipeline presented in [17] highlights the AP features. Our method uses these AP features as a guide for generating masks for inpainting with human supervision. Using the mask of road view images made by a human operator for accident hotspot locations, images are inpainted by a diffusion model for enhancing the safety of the scene. The overall road redesign process of our novel method is shown in Fig. 1. Firstly, using the dataset of actual accident events, both hotspots and non-hotspots are collected. A hotspot is identified based on clustering locations of accident events and by using the location road view images of hotspots and non-hotspots are then collected. From this context, a hotspot is an area rather than a single location. A deep learning classifier is trained to detect hotspot images and then various types of CAM [17] such as GradCAM, GradCAM++, and ScoreCAM can be leveraged to inspect what AP features lead to the classification of a hotspot. From these AP features, masks, which will be inpainted, are generated by human-in-the-loop. The AP features mask can be combined with other masks of road markings, traffic signs, and traffic signals to make a saliency mask. Safety critical elements such as road markings that might not be detected by CAM due to their common presence in both hotspots and non-hotspots should also be considered to achieve road safety. A small dataset of safe-road design features is collected and used to fine-tune the diffusion model with a text prompt, such as a class prompt, along with a subject word. For this work, the seven safe road elements listed in TABLE II are used for our diffusion model. The base diffusion model used for this work is Hugging Face stable-diffusion-v1-5 and fine-tuning is executed with DreamBooth [23] and Textual inversion [24]. For the saliency mask designated by the human operator, safe road elements also chosen by a human operator are inpainted with the fine-tuned diffusion model. Because the color of the generated safe road elements can be close to that of adjoining roadway parts, chrominance alteration might be necessary. To this end, a saliency model such as GazeShiftNet can be used.

Features of this article can be listed as follows

- An image inpainting technique that can assist authorities in achieving safe roadway design with minimal intervention in the current structure of a roadway is introduced.
- Demarcation of AP features in street view images of accident hotspots is presented.
- With the fine-tuning of a diffusion model, a methodology for safe road design ensuring minimal intervention of current road design is introduced.
- For redirecting a driver's attention towards AP features and other accident critical elements in road view scenes, visual saliency enhancement by chrominance alteration is presented.

Each feature is fully explained in Section IV.

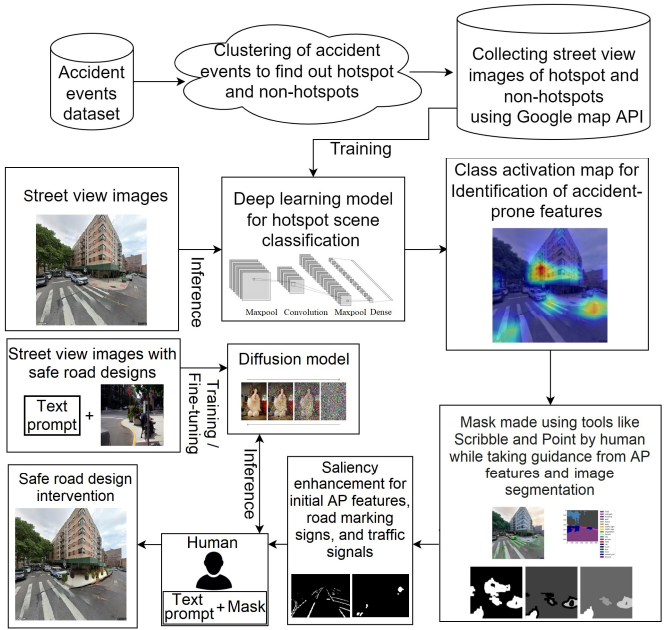

**Fig. 1.** Road redesign process in our methodology.

In [16], GazeShiftNet is presented, which is a model to enhance the saliency of important safe design features in images for a given (saliency) mask while preserving image fidelity. GazeShiftNet can be used to enhance the saliency by chrominance alteration for changing color(s) of AP features as well as marking signs and traffic signals with bright color(s). When the bright colors obtained from the use of the model are actually used, they can redirect the drivers' attention for improved driving safety. In [17], a visual notification system is presented using class activation maps (CAMs) to activate only the important AP features for the classification of

## II. RELATED WORKS

### A. AP features

AP features are the unsafe features of a road in the visual area of the driver's view and are significantly related to

# Replace this line with your manuscript ID number (double-click here to edit)

3accident occurrence. These unsafe features in the road view have been investigated in previously published works. Some works use a manual inspection process while others use an automatic process by leveraging machine learning techniques. In [18], a cognitive work analysis was used for measuring the effectiveness of road design features based on safety, positive subjective experience, and compliance by drivers. Similarly, in [19], the impact of design features such as zebra crossings, speed bumps, etc., which make roads safe and self-explanatory, were studied.

Machine learning techniques like regression trees have been used to classify road intersections associated with vehicle-to-pedestrian collisions. In [22], convolutional neural networks (CNNs) are used to analyze satellite images of intersections after extracting high-level features by using an autoencoder. These features are clustered in an unsupervised way based on accident events. This AP feature detection method provides a reliable and objective assessment of road design features. However, the unsupervised method still can not adequately discriminate AP features, but points out the features for being accident-prone. Therefore, as a straightforward methodology, supervised detection of AP features can be used. In our work, a historical accident dataset is used for identifying accident hotspots. Street view images of those hotspot locations are collected to train a binary classification model. For a more effective and direct search of AP features, a CAM-based method is leveraged to inspect why a particular street view image was chosen as a hotspot image.

*B. Safe road design*

Road design engineering focuses on factors that mitigate the risk of accidents due to speeding, blind curves, vehicle-pedestrian collision, etc. [18]. A safe structural design of pathways reduces accident proneness. Road design features such as chicane, choker, and roundabouts help in creating safer and more efficient environments by encouraging drivers to lower driving speed. Chicanes are a series of alternating mid-block curb extensions that narrow the roadway and require vehicles to follow a curving (S-shaped) path, whereas chokers are curb extensions that narrow a street by widening the side-walks or planting strips, effectively creating a pinch point along the street. Roundabouts in large intersections help in both reducing speed and organizing traffic. Raised medians are barriers in the center portion of a street or roadway, helping in speed reduction as well as providing refuge for pedestrians crossing the road. In addition to limiting vehicle speed, to reduce the vehicle-pedestrian collision rate [21], treatment of curb extensions, medians and street plazas is instrumental. Street plazas are semi-enclosed pedestrian-friendly zones adjoining a sidewalk or transit stop while curb extensions widen a sidewalk for a short distance creating safer crossings for pedestrians.

*C. Fine-tuning technique of diffusion model*

Recent advances in the latent variable based deep generative model such as the diffusion model and growing interest in personalizing the final output image have led to the study of fine-tuning with a small set of personalized data. DreamBooth, an approach to fine-tuning diffusion models, is presented in [23]. In this approach, a new subject is embedded in the output domain of the model using a new loss function named as autogenous class-specific prior preservation loss. Therefore, the newly added subject can be contextualized in different scenes to generate photorealistic images. Textual inversion [24] is another approach that uses new 'words' (guiding personalized creation) in the embedding space of pre-trained text-to-image models. Pre-trained models enhance the picture quality manifold as compared to from-scratch techniques, generating high resolution and aesthetically appealing realistic results. The developer community tested and compared these two methods [26], and a combination of both has shown improved results [27]. Therefore, we employ both fine-tuning techniques for the diffusion model to introduce design modifications for safer streets and roadways. Firstly, the diffusion model is fine-tuned using the DreamBooth technique and then Textual inversion is used with the fine-tuned model. Inpainting with the diffusion model requires an image, mask, and text as inputs. The mask is created manually while being guided by AP features as well as object segmentation.

### III. Data Construction

For AP feature detection, two datasets are needed: (a) a dataset of real accident events and (b) image data of accident hotspots. For (a), we used real accident event data provided by New York City [17]. The DBSCAN algorithm, widely used for clustering, was chosen for the identification of accident hotspots because of its proven efficacy in [29], [30], and [31]. The average latitude and longitude location of all the accidents of a given cluster is used to find the center of the hotspot. The algorithm in [28] detects hotspots based on raw vehicle data such as vehicle braking, accelerating, and frequency of accidents. For (b), using Google street view, we capture images for the center location of hotspots to cover an approximately 240 degree field of view. Accordingly, 5,088 images belonging to hotspots were collected. Similarly, 4,908 street view images of non-hotspots were randomly collected to make a balanced dataset for classifier training.

As per previous studies on roadway design engineering for safe streets [32], [33] and minimizing accident risks [34], [18], [21], safe road structure plays a major role; therefore, for this work, seven major safe road designs were finalized: chicanes, chokers, curb extensions, raised medians, roundabouts, street plazas, and road markings on big intersections. Six to seven street image samples of each safe road design were curated to the pre-train diffusion models.

### IV. Proposed Mechanism

*A. Selection of AP features*

In our model, the AP features are extracted by post-hoc methods such as the CAM-based method. The CAM-based method provides information by a deep learning classifier specifically about the region of the image that contributes most to classifying that image as a hotspot. For the deep learning classifier, CNN backbones, as shown in Fig. 2, are leveraged. Along with a CNN backbone, a fully connected (FC) layer with two outputs acting as a classifier is used with the Softmax normalizing function to obtain the class probability. In [17], an attention based module (ABM) to improve the inherent interpretability of the CNN architecture and select more contextual AP features is proposed. The ABM generates the attention maps and attention vectors in the training process to appropriately weigh the spatial, channel, and point features from the CNN backbone. This attention based activation can characterize the target area with improved context.

For AP feature detection, CNN backbones like Squeezenet, Resnet, VGG, and Densenet can be used for classification. TABLE I shows the metrics for classification for 30% of randomly selected test images out of the 9,996 road view images. Accuracy is the indicator of the ratio of correct prediction to the total number of input samples. Precision indicates the ratio of a total number of correct prediction results to the positive results as predicted by the classification model. Recall indicates the ratio of a correct positive result by the classification model to the total number of positive samples. The F1 score is the indicator of precision as well as the robustness of the model and is the mathematical harmonic mean of the precision and recall. The accuracy metric with the ABM block is better for most CNN backbones. Densenet, as a backbone with an ABM module, provides the most accurate results. To detect AP features, different types of CAM such as GradCAM, GradCAM++, and ScoreCAM are used to inspect features leading to the classification of a hotspot. GradCAM, when applied to Squeezenet-ABM for hotspot classification, gives best AP features having more context, as stated in [17].

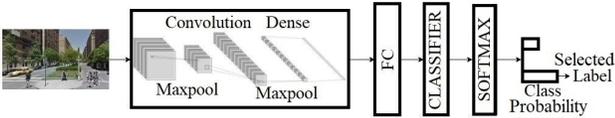

**Fig. 2.** Generic layer-wise architecture of the deep learning model for classification of hotspot and non-hotspot images. The large box represents the CNN backbone.

TABLE I
ABLATION STUDY FOR ABM ALONG WITH DIFFERENT CNN BACKBONES (✘ REPRESENTS ABM NOT INCLUDED AND ✓ REPRESENTS ABM INCLUDED).

| CNN Back-bone | ABM Block | Accuracy | Precision | Recall | F1-Score |
|---|---|---|---|---|---|
| Squeezenet | ✘ | 0.891 | 0.902 | 0.859 | 0.864 |
|  | ✓ | 0.892 | 0.898 | 0.876 | 0.871 |
| Resnet | ✘ | 0.900 | 0.912 | 0.892 | 0.889 |
|  | ✓ | 0.900 | 0.915 | 0.869 | 0.875 |
| VGG | ✘ | 0.904 | 0.916 | 0.888 | 0.887 |
|  | ✓ | 0.902 | 0.926 | 0.867 | 0.880 |
| Densenet | ✘ | 0.905 | 0.903 | 0.876 | 0.877 |
|  | ✓ | 0.922 | 0.947 | 0.901 | 0.913 |

*B. Mask generation for inpainting by human-in-the-loop*

The introduction of safe road design with minimal intervention in the current structure of a roadway requires street view images and corresponding masks as input for inpainting. However, generating a mask for inpainting a safe design with minimal intervention is complex. For street images, making masks using only AP features may be vague, because the features may not enclose quantifiable objects. For example, masks for sidewalks can have a broader proportionate area than the actual sidewalk, including sections of the adjoining road area. The masks for greenery on the roadside should be touching the ground and can be of some height alongside the road for inpainting bushes or trees accordingly.

For simplicity, masks can be generated by a deep learning based object segmentation model with human interaction such as clicks [35], scribbles [36], bounding boxes [37], [38], [39], or extreme point selections [40]. Figure 3 (a) presents object segmentation and Fig.3 (b) provides AP features detected by the CAM-based method. A human operator generates masks using the green marking, as shown in Fig. 3 (c). Object segmentation can be used while considering AP features so that, if possible, the meaningful objects are considered while creating masks [35]. Scribbles is best suited [41] for complex images such as street view images. For our purpose, a human operator can add strokes for mask making while considering road redesign construction factors such as the required time and type of construction, funds, ease of deployment, traffic safety, etc. After mask creation, for the generation of inpainted images based on the provided mask and street view image, the fine-tuned diffusion model can be used. Fine-tuning of the diffusion model is done on seven road design components: chicanes, chokers, curb extensions, raised median, roundabouts, street plazas, and road markings on large intersections.

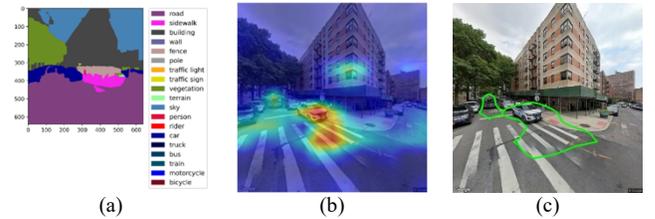

**Fig. 3.** Guidance for mask generation for inpainting: (a) object segmentation; (b) heatmap of AP features; (c) marking of driver relevant AP features in green.

*C. Training and inferencing*

After detection, we then generate a safe road design to replace the AP features. The region considered for safe road design generation is marked by a mask, as described in the previous sub-section. For this purpose, a diffusion model which is already trained on generic data is further trained or fine-tuned to generate a specific safe road design. Fine-tuning of the diffusion model is done using DreamBooth and Textual inversion. The base diffusion model used is Hugging Face stable-diffusion-v1-5. For DreamBooth, class prompts describing the property of safe road designs are used to generate some example images for generalization while training. Class prompts describe a new design, but in generic English language that has been used for base model training. For this purpose, respective text representing class prompts for each road design, as shown in TABLE II, are used. To generalize, 50 images for each class are generated. As the text based inpainting model is fine-tuned to create new safe road designs, new textual names for those safe road designs are needed during training and inferencing. These texts are called subject words. In the class prompt text, as a new subject, a random word not in the English language is preferably used as an identifier word to bind a unique identifier with that specific safe road design. For DreamBooth, a new subject word in the class prompt after the words "photo of …" was included. Training was done for 2000 epochs with a learning rate of '$1\times10^{-6}$'. The original hyperparameters, as used in [23], are adopted. The training script, suitable for Google Colab, is available in [42].

For Textual inversion, a model trained by DreamBooth is used. The same subject word is used to create embedding that has been used for DreamBooth for a particular safe road design. To create embedding, 8 tokens per word are selected. The class prompts are used and added in the text file for each respective image for input in the training instance, as provided in the Automatic1111 UI training tab [43]. When training the embeddings of Textual inversion, some text describing the image along with the class prompt and new subject is provided for each image in a text file template. This text prompt is used to create similar images for generalization. Training is performed for 2000 epochs with an embedding learning rate of 0.005. Other hyperparameters are similar to that of the original setting in [24] and is pre-set in the Automatic1111 UI training session.

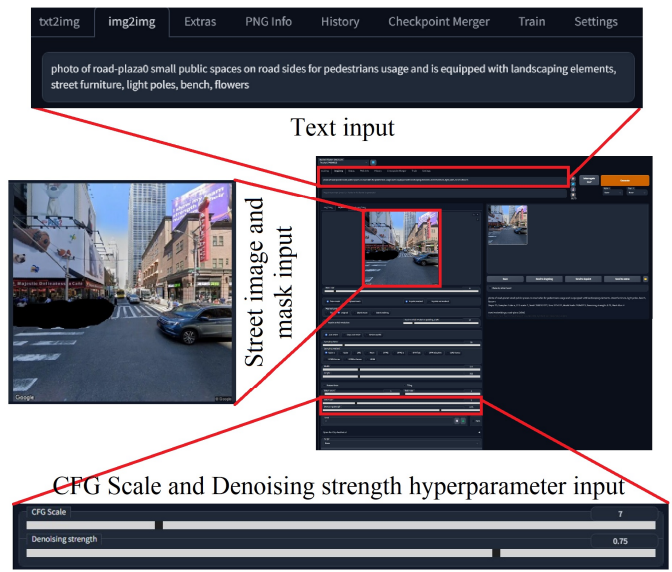

Fig. 4. Automatic1111 user interface dashboard shown for inferencing along with different inputs.

Following training, inferencing for new safe road design is conducted. For efficient inferencing, Automatic1111 is widely used. Figure 4 shows the user interface (UI) of Automatic1111 along with different inputs and hyperparameters. The trained DreamBooth model or the fine-tuned model is loaded in the Automatic1111 UI. For inpainting, a mask is required which can be drawn directly in Automatic1111 or can be a separate input. Text input is also required to guide the generation of a new inpainted image. For the text input (prompt), the class prompt along with a new subject word, as used in DreamBooth training, is used. The text should be given according to the position of the mask and the requirement of creating a particular safe road design. For inpainting, other various hyperparameters can be calibrated with the UI. For example, the sampling method can be chosen from a range of available options. The classifier-free guidance scale (CFG Scale), responsible for the weight dependency of the text prompt on the inpainted image, can be adjusted from a range of 0 to 30. Similarly, Denoising strength provides the weight dependency of the inpainted region to the original image. This can also be adjusted by a slider in the UI and varies from 0 to 1. After generating a few samples, a human operator can select the best one. From our empirical experience, a suitable range of 7 to 18 for the CFG Scale and a range of 0.65 to 0.75 for Denoising strength is best for designing photorealistic results.

TABLE II
SAFE ROAD DESIGN STRUCTURES AND THEIR NEW SUBJECT WORD AND CLASS PROMPT AS USED FOR DIFFUSION MODEL FINE-TUNING.

| Safe road design structure | New subject word in class prompt | Class Prompt |
|---|---|---|
| Chicane | road-chicane0 | hoto of S-shaped curve in the vehicle driving path, created by offset curb extensions in straight road |
| Choker | road-choker0 | photo of parallel or offsetting curb extensions, which effectively reduce road width for a specific distance |
| Curb Extension | road-curb0 | photo of extension of sidewalk at intersection for reducing crossing distance and increasing visibility |
| Raised Median | road-median0 | photo of barriers in center portion of street or roadway separating different lanes and traffic direction |
| Roundabout | road-circle0 | photo of roundabouts or traffic circle with a circular central space in middle of an intersection |
| Street Plaza | road-plaza0 | photo of small public spaces on road sides for pedestrians usage and is equipped with landscaping elements, street furniture, light poles, bench, flowers |
| Big Intersection | road-intersection0 | photo of big intersection with bus corridors, different lane marking, crossways |

*D. Visual Saliency for AP features*

To find the visual complexity, [11] illustrates the concept of Scene Disorder (SD), directly related to the number of object categories present in a given scene. A higher SD value implies a more complex image resulting in reduced attention towards objects that are relevant to accident risk. However, SD might not be directly linked to driver attention. As a more direct metric, the saliency drawing a driver's attention in complex scenes can be compared with the AP features of the scene. If the saliency of the AP features is less, indicating a scene is complex, then the driver may divert their attention away from the features making them more prone to accidents. As a metric, saliency is defined as the percentage ratio of the salient AP feature to the whole AP feature area. A salient area of the original image detected by the GazeShiftNet [16] is calculated and its intersection with an AP feature detected by the CAM-based method is defined as the salient AP feature and calculated for given images. Then, the percentage ratio of the intersected area with respect to the AP feature area is calculated. The average of the percentage ratio is taken for all images and enlisted, as shown in Table III. Here, we studied different architectures along with different CAMs. As a result, we found that a combination of Squeezenet-ABM with GradCAM gives the lowest value of visual saliency. This shows that the best AP features have the least saliency, and to make the road view safe, some intervention in the form of chrominance alteration is required for design improvement. This will increase the saliency of AP features by introducing contrastive bright colors to make them more noticeable.

TABLE III
VISUAL SALIENCY FOR DIFFERENT MODELS AND CAMS. THE LOWEST VALUE IS HIGHLIGHTED IN BOLD.

| Model | GradCAM | GradCAM++ | ScoreCAM |
|---|---|---|---|
| VGG-ABM | 36.92 | 41.34 | 24.71 |
| Resnet-18-ABM | 46.59S | 48.61 | 44.98 |
| Squeezenet-ABM | **23.98** | 27.46 | 31.42 |
| Densenet-ABM | 40.23 | 53.38 | 47.31 |

*E. Saliency enhancement*

The physical saliency of items that attract attention is a prominent factor affecting a driver's behavior [19]. Therefore, saliency enhancement of AP features, road markings, signs, and traffic signals in the road view scene of drivers can increase safety. For this, masks are generated based on object segmentation and AP feature maps created by the use of a CAM-based method. Object segmentation helps identify different categories of objects present in the scene. For example, traffic signals and signs are detected. For road marking, we leverage the Inter-Region Affinity KD method presented in [44]. After combining all the masks of the AP feature maps, road markings, traffic signs, and signals, the saliency mask is generated.

For a given image and saliency mask, the saliency model is used to change the saliency in the masked region [51]. This results in a generated image that redirects the attention of a driver to the AP features. The new inpainted image of safe structures and the final mask generated by using the original road view image are used as the two inputs into the deep saliency model.

V. OBSERVATIONS AND EXPERIMENTS

*A. Analysis of road design intervention*

As mentioned in the inferencing stage in Subsection IV-C, a human operator is present to decide the mask and text prompt for inpainting. The operator can also adjust the input hyperparameters and select the best visually inpainted result. Using the ABM based deep learning classifier, as used for selection of AP features, we selected 50 images classified as hotspot images. The mask, text prompt, hyperparameter setting, and selection of the best result is performed by a human operator to obtain a new safe road element with minimal intervention for each of the 50 images. Samples of the safe road elements are shown in Fig. 5. A human operator spent approximately 2 min, on average, for each image. From the various results, as shown for the case in Fig. 6, cherry picking of the results can be done according to the time and type of construction required, funds, ease of deployment, traffic safety, etc. The qualitative results show that with minimalistic changes, a safe road design can be introduced to enhance the safety of the road view.

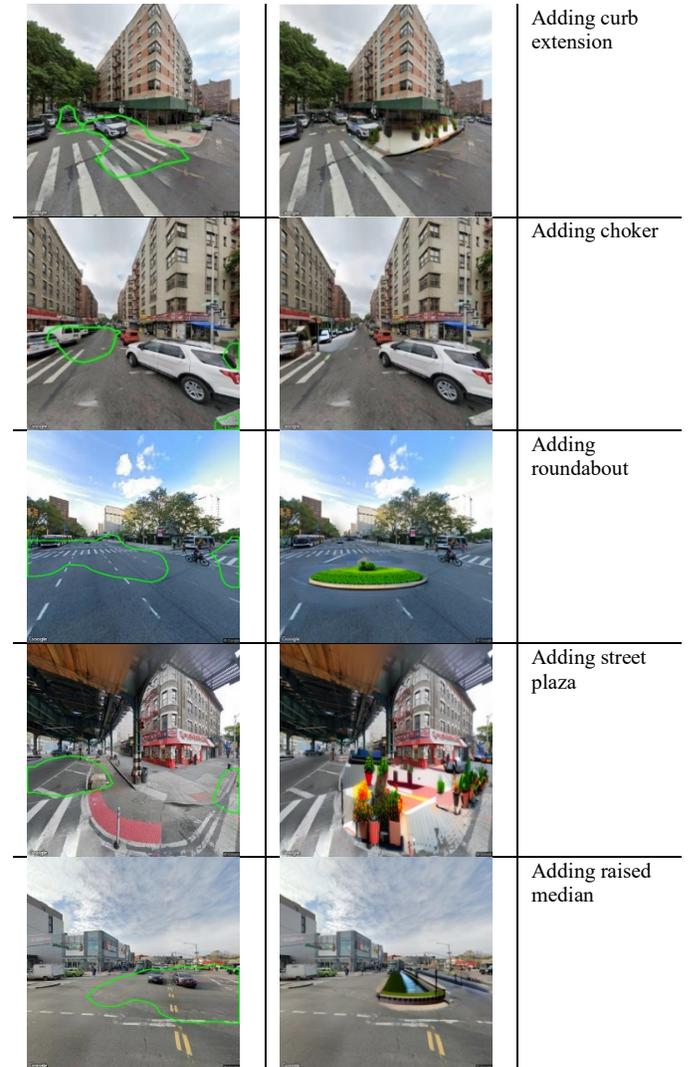

**Fig. 5.** Road view inpainted with safe road elements. Left column represents the original images and middle column shows inpainted images and right column presents the names of newly added safe road elements.

Furthermore, as a quantitative study, we note the average probability of the deep learning classifier to select 50 images as hotspot images. Different CNN backbones along with the ABM module were tested. The results of the qualitative study for the 50 images are shown in TABLE IV. The hotspot classification probability of the best model, Squeezenet-ABM with the new safe and intervened images, dropped by 11.85% on average. Even though the percentage drop by the Densenet CNN backbone is larger, it is not considered as a suitable model, as its average classification probability for the original 50 images is just 0.70.

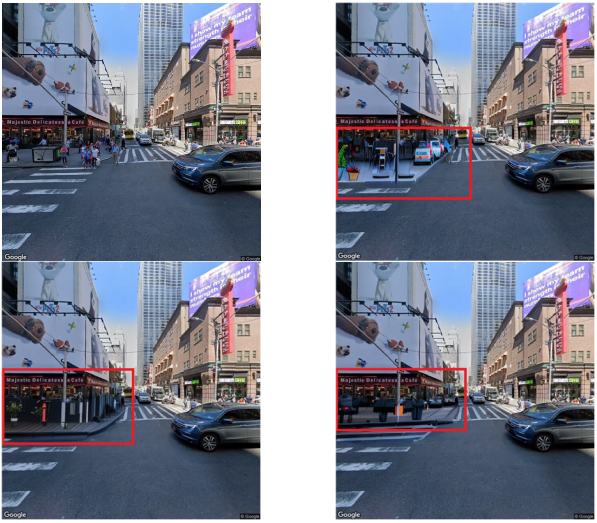

**Fig. 6.** Different results of inpainting (marked with a red box) with safe road elements for the given image in the top-left.

TABLE IV
STUDY FOR HOTSPOT CLASSIFICATION BY DEEP LEARNING CLASSIFIER OF 50 HOTSPOT IMAGES AND THEIR RESPECTIVE SAFE DESIGN INPAINTED IMAGES.

| CNN backbone (ABM is present on top of CNN backbone) | Avg. probability of classification into hotspot class with 50 original scenes | Avg. probability of classification into hotspot class with 50 inpainted scenes | Percentage Change |
|---|---|---|---|
| VGG-ABM | 0.98 | 0.95 | 3.2 |
| Densenet-ABM | 0.70 | 0.59 | 15.99 |
| Resnet-18-ABM | 0.99 | 0.91 | 8.45 |
| Squeezenet-ABM | 0.97 | 0.86 | 11.85 |

*B. Discussion on enhancing Visual saliency for new road design intervention*

As stated, saliency enhancement by chrominance alteration can mitigate accident risk by introducing bright contrastive color in the saliency mask area. Therefore, we performed experiments to increase the saliency of AP features, road signs, road markings, and traffic signals in road view images by using the saliency mask. Observing the new road view scenes with enhanced saliency, frequent road management practices such as timely paint coating of traffic signals, road marking, and road signs are suggested. As per safe road designs [46], authorities can undertake measures such as raised crossings, speed bumps, etc. to increase visibility of roads wherever necessary. Moreover, other road design interventions for enhancing visual saliency can use contrast color to highlight potential conflict zones or intersecting areas of lanes [47]. For this, the use of photoluminescent paint for pavements to increase night-time visibility [48], colored asphalt pavements durable in dry and wet weather conditions [49], and retroreflective material-based road signs and markings [50] should be encouraged.

## V. CONCLUSION

In this article, a methodology for minimal intervention in current road design under human supervision to mitigate accident risk is proposed. This road redesign methodology is introduced considering two levels: safe road design and the human factor of road users. With our methodology, street images are first classified as accident hotspots by using a deep neural network. The CNN backbone Densenet along with ABM shows 92% accuracy and 94% precision. For a more precise contextual selection of AP features, we also inspected SqueezeNet-ABM using GradCAM. This classification helps determine the AP features that are the major cause for classification of a given image into an accident hotspot. Road design elements that are easy to deploy as well as impactful in preventing accidents are explored. Using this, we introduce a methodology for safe road design using fine-tuning techniques for a diffusion model such as DreamBooth and Textual inversion. Under human supervision, inpainted street images with safe design features such as a raised median, roundabout, etc. are generated. The image inpainted with a safe road design reduces the chance of hotspot classification by approximately 11.85% with SqueezeNet-ABM. Additionally, to understand scene complexity, we discussed the impact of the overlap of the mask area between the saliency area and AP features mask. Our assessment shows that, in complex street view scenes, saliency of AP features is subdued. Thus, visual saliency enhancement for making AP features with bright and contrastive colors is required as an intervention to attract a driver's attention. For this, chrominance alteration by using contrasted painting of pavements and speed bumps, use of retroreflective material for road markings and signages, etc. are suggested. Chrominance alteration is likely to redirect a driver's attention to AP features, road markings, signs and traffic signals, thereby aiding in accident prevention.

## VI. ACKNOWLEDGMENT

This work was supported in part by the Institute of Information and Communications Technology Planning and Evaluation (IITP) Grant funded by the Korea Government (MSIT) (Development of Artificial Intelligence Technology that Continuously Improves Itself as the Situation Changes in the Real World) under Grant 2020-000440.

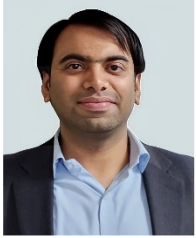
**Sumit Mishra** received B. Tech. in Electronics and Communication Engineering from Dr. A.P.J. Abdul Kalam Technical University, India, in 2016, and M.S. degree from Robotics department, Korea Advanced Institute of Science and Technology, South Korea, in 2023, where he is currently pursuing the Ph.D. degree. Earlier, he worked with Learnogether Technologies Pvt. Ltd. as Research Consultant. His research interest is in deep learning, computer vision, and robotics perception using multi-modal data.

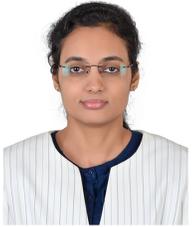
**Medhavi Mishra** received B. Tech. in Production and Industrial Engineering from Motilal Nehru National Institute Of Technology, Allahabad, India, in 2016. She is pursuing M.S. degree in Green Mobility from Korea Advanced Institute of Science and Technology, South Korea. Earlier, she worked with HighRadius Technologies Pvt. Ltd. as Techno-functional Consultant. Her research interest is in optimisation technique application, crowd management and pedestrian flow simulation.

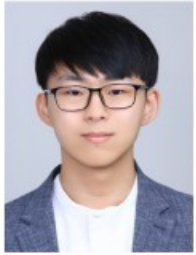
**Taeyoung Kim** received the B.S. degree in electronic and electrical engineering from the Sungkyunkwan University, South Korea, in 2019, and the M.S. degree from The Cho Chun Shik Graduate School of Mobility, Korea Advanced Institute of Science and Technology, South Korea, where he is currently pursuing the Ph.D. degree. His main research interests include the areas of machine learning, reinforcement learning, and multi-goal reinforcement learning.

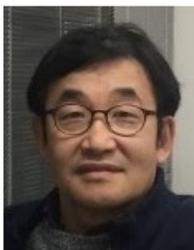
**Dongsoo Har** (Senior Member, IEEE) received the B.Sc. and M.Sc. degrees in electronics engineering from Seoul National University, and the Ph.D. degree in electrical engineering from Polytechnic University, Brooklyn, NY, USA. He is currently a Faculty Member of KAIST. He has authored and published more than 100 articles in international journals and conferences. He has also presented invited talks and keynote in international conferences. His main research interests include optimization of communication system operation and transportation system development with embedded artificial intelligence. He was a member of the Advisory Board, the Program Chair, the Vice Chair, and the General Chair of international conferences. He was a recipient of the Best Paper Award (Jack Neubauer Award) from the IEEE TRANSACTIONS ON VEHICULAR TECHNOLOGY in 2000. He is an Associate Editor of the IEEE SENSORS JOURNAL.